%% file: main.tex
\documentclass[sigconf]{acmart}  
\AtBeginDocument{%
  \providecommand\BibTeX{{%
    \normalfont B\kern-0.5em{\scshape i\kern-0.25em b}\kern-0.8em\TeX}}}

\setcopyright{acmcopyright}
\copyrightyear{2022}
\acmYear{2022}
\acmDOI{}

\acmConference[KDD '22]{KDD '22: ACM Symposium on Neural Gaze Detection}{August 14--18, 2022}{Washington, DC, USA}
\acmBooktitle{Proceedings of the 28th ACM SIGKDD Conference on Knowledge Discovery and Data Mining (KDD ’22), August 14--18, 2022, Washington, DC, USA}
\acmPrice{}
\acmISBN{}
\usepackage{subfig}
\usepackage{algorithm}
\usepackage{algorithmicx}
\usepackage{enumitem}

\newcommand{\med}{\mbox{GraphAKD}}

\usepackage{algpseudocodex}
\usepackage{multirow}
\usepackage{graphicx} 
\usepackage{xcolor}

\begin{document}

\title[Compressing Deep Graph Neural Networks via Adversarial Knowledge Distillation]{Compressing Deep Graph Neural Networks via\\ Adversarial Knowledge Distillation}



\author{Huarui He}
\orcid{0000-0002-2905-4331}
\affiliation{%
  \institution{CAS Key Laboratory of Technology in GIPAS, \\
University of Science and Technology of China}
  \streetaddress{No. 666 Innovation Avenue}
  \city{Hefei}
  \country{China}}
\email{huaruihe@mail.ustc.edu.cn}

\author{Jie Wang}
\authornote{Corresponding Author.}
\email{jiewangx@ustc.edu.cn}
\affiliation{%
  \institution{CAS Key Laboratory of Technology in GIPAS, \\
University of Science and Technology of China}
  \institution{Institute of Artificial Intelligence, \\
Hefei Comprehensive National Science Center}
  \streetaddress{443  Huangshan Rd}
  \city{Hefei}
  \country{China}
  \postcode{230027}
}

\author{Zhanqiu Zhang}
\affiliation{%
  \institution{CAS Key Laboratory of Technology in GIPAS, \\
University of Science and Technology of China}
  \streetaddress{No. 666 Innovation Avenue}
  \city{Hefei}
  \country{China}}
\email{zqzhang@mail.ustc.edu.cn}

\author{Feng Wu}
\affiliation{%
  \institution{CAS Key Laboratory of Technology in GIPAS, \\
University of Science and Technology of China}
  \streetaddress{No. 666 Innovation Avenue}
  \city{Hefei}
  \country{China}}
\email{fengwu@ustc.edu.cn}



%
\renewcommand{\shortauthors}{He et al.}

\begin{abstract}
  \input{section/abstract}
\end{abstract}



\maketitle
\input{section/intro}

\input{section/background}

\input{section/med}
\input{section/exp}
\input{section/related}

\input{section/conclusion}

\bibliographystyle{ACM-Reference-Format}
\bibliography{kdd2022}

\appendix
\input{section/appendix}

\end{document}

%% file: section/abstract.tex
Deep graph neural networks (GNNs) have been shown to be expressive for modeling graph-structured data.
Nevertheless, the over-stacked architecture of deep graph models makes it difficult to deploy and rapidly test on mobile or embedded systems. 
To compress over-stacked GNNs, knowledge distillation via a teacher-student architecture turns out to be an effective technique, where the key step is to measure the discrepancy between teacher and student networks with predefined distance functions.
However, using the same distance for graphs of various structures may be unfit, and the optimal distance formulation is hard to determine.
To tackle these problems, we propose a novel Adversarial Knowledge Distillation framework for graph models named \med, which adversarially trains a discriminator and a generator to adaptively detect and decrease the discrepancy.
Specifically, noticing that the well-captured inter-node and inter-class correlations favor the success of deep GNNs,
we propose to criticize the inherited knowledge from node-level and class-level views with a trainable discriminator.
The discriminator distinguishes between teacher knowledge and what the student inherits,
while the student GNN works as a generator and aims to fool the discriminator.
To our best knowledge, \med~is the first to introduce adversarial training to knowledge distillation in graph domains.
Experiments on node-level and graph-level classification benchmarks demonstrate that \med~improves the student performance by a large margin. The results imply that \med~can precisely transfer knowledge from a complicated teacher GNN to a compact student GNN.

%% file: section/intro.tex
\section{Introduction}
\begin{figure}[!t]
  \centering
  \includegraphics[width=\linewidth]{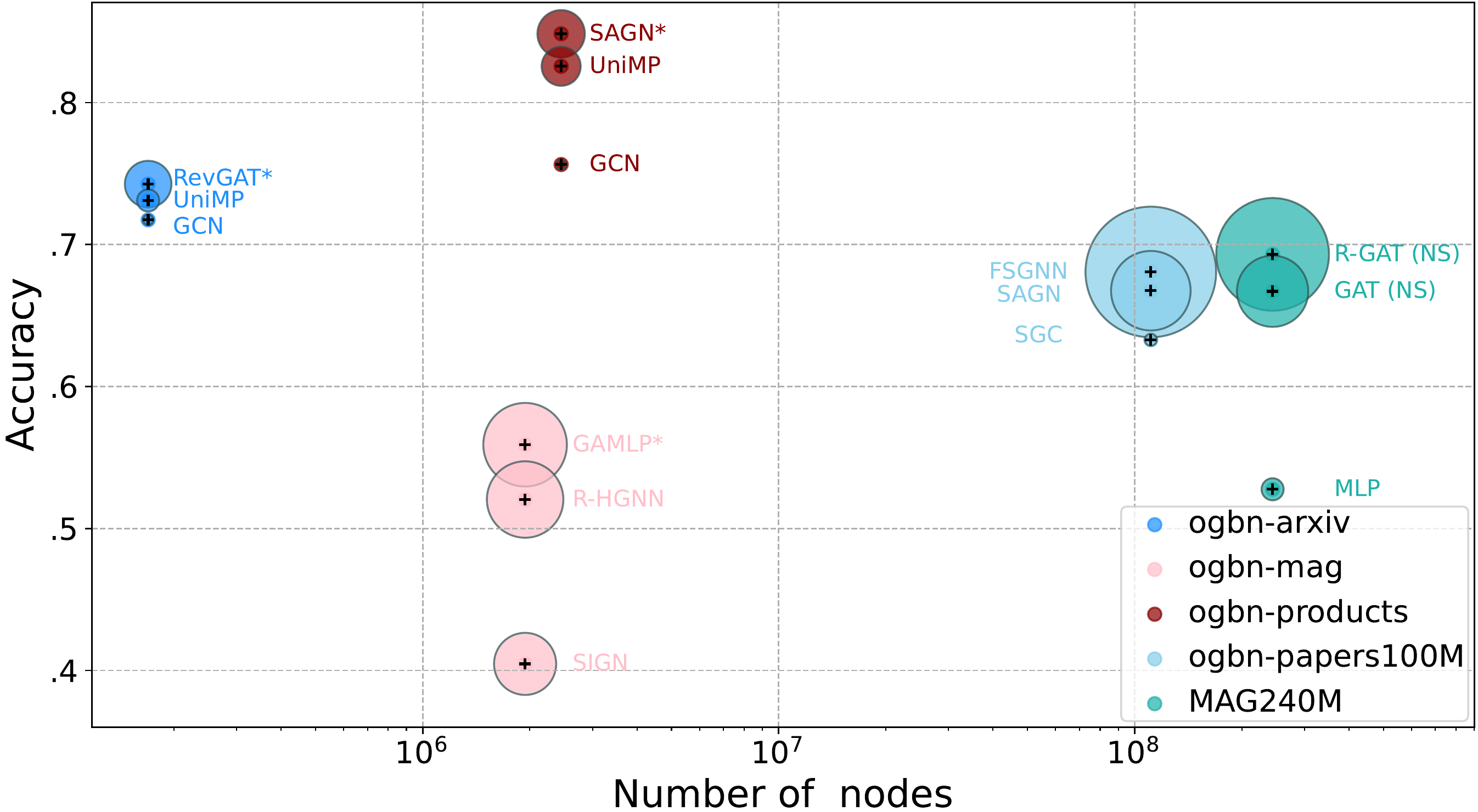}
  \caption{Node classification accuracy v.s. graph size. Each bubble's area is proportional to the number of parameters of a model. Model name with * means the variant. The statistics are collected from OGB leaderboards.}
  \label{fig:bubble}
  \vspace{-4ex}
\end{figure}

In recent years, graph neural networks (GNNs) have become the standard toolkit for graph-related applications including recommender systems \cite{DBLP:conf/kdd/YingHCEHL18,lightgcn}, social network \cite{reddit,DBLP:conf/acl/LiG19}, and biochemistry \cite{DBLP:conf/nips/DuvenaudMABHAA15,DBLP:conf/nips/FoutBSB17}.
However, GNNs that show great expressive power on large-scale graphs tend to be over-parameterized \cite{ogb-lsc}.
As large-scale graph benchmarks including Microsoft Academic Graph (MAG) \cite{mag} and Open Graph Benchmark (OGB) \cite{ogb} spring up, complicated and over-stacked GNNs \cite{gcn2,revgnn} have been developed to achieve state-of-the-art performance.
Figure \ref{fig:bubble} illustrates model performance versus graph size (i.e., the number of nodes in a graph).
We note that deep and complicated GNNs significantly outperform shallow models on large-scale graphs, implying the great expressive power of over-parameterized GNNs.
However, the over-stacked architecture frequently and inevitably degrades both parameter-efficiency and time-efficiency of GNNs \cite{revgnn}, which makes them inapplicable to computationally limited platforms such as mobile or embedded systems.
To compress deep GNNs and preserve their expressive power, we explore the knowledge distillation technique in graph domains, which has attracted growing attention in recent years.

Knowledge distillation has been shown to be powerful for compressing huge neural networks in both visual learning and language modeling tasks \cite{DBLP:conf/cvpr/BergmannFSS20,tinybert}, especially with a teacher-student architecture.
The main idea is that the student network mimics the behavior of the teacher network to obtain a competitive or even a superior performance \cite{ban,fitnets},
while the teacher network transfers soft targets, hidden feature maps, or relations between pair of layers as distilled knowledge to the shallow student network.
However, existing algorithms that adapt knowledge distillation to graph domains \cite{DBLP:conf/cvpr/YangQSTW20,DBLP:conf/sigmod/ZhangMSJCR020,DBLP:conf/www/0002LS21} mainly propose specially designed and fixed distance functions to measure the discrepancy between teacher and student graph models, which results in following two inherent limitations.
\begin{itemize}[leftmargin=*]
    \item They force the student network to mimic the teacher network with hand-crafted distance functions, of which the optimal formulation is hard to determine  \cite{DBLP:conf/aaai/WangXXT18}. Even worse, \citet{kdgan, DBLP:journals/pami/WangZSQ21} have pointed out that the performance of the student trained this way is always suboptimal because it is difficult to learn the exact distribution from the teacher.
    \item The predefined and fixed distance is unfit to measure the distribution discrepancy of teacher and student representations in different feature spaces. For example, citation networks and image networks have distinct feature spaces due to the intrinsic difference between textual data and visual data. Experiments in Section \ref{sec:node-level results} also confirm this claim.
\end{itemize}

In this paper, we propose a novel adversarial knowledge distillation framework named \med~to tackle the aforementioned problems.
Specifically, instead of forcing the student network to exactly mimic the teacher network with hand-designed distance functions,
we develop a trainable discriminator to distinguish between student and teacher from the views of node representations and logits.
The discriminator modifies the teacher-student architecture into generative adversarial networks (GANs) \cite{gan}, where the student model works as a generator.
Two identifiers constitute the discriminator, namely the represenation identifier and the logit identifier.
The representation identifier tells student and teacher node representations apart via criticizing the local affinity of connected nodes and the global affinity of patch-summary pairs,
while the logit identifier distinguishes between teacher and student logits with a residual multi-layer perceptron (MLP).
{We think the proposed discriminator is topology-aware as it considers graph structures.}
The generator, i.e., the student network, is trained to produce node representations and logits similar to the teacher's distributions so that the discriminator cannot distinguish.
By alternately optimizing the discriminator and the generator, \med~is able to transfer both inter-node and inter-class correlations from a complicated teacher GNN to a compact student GNN.
We further note that the discriminator is more tolerant than predefined distance formulations such as Kullback-Leibler (KL) divergence and Euclidean distance.
We can view the trainable discriminator as a teaching assistant that helps bridge the capacity gap between teacher and student models.
Moreover, the proposed {topology-aware} discriminator can lower the risk of over-fitting.

To evaluate the effectiveness of our proposed \med, we conduct extensive experiments on eight node classification benchmarks and two graph classification benchmarks.
Experiments demonstrate that \med~enables compact student GNNs to achieve satisfying results on par with or sometimes even superior to those of the deep teacher models, while requiring only 10 $\sim$ 40\% parameters of their corresponding teachers.


%% file: section/background.tex
\begin{table}[!t]
\caption{Glossary of notations.}\label{tab:notation}
\vspace{-1.5ex}
    \centering
    \resizebox{\columnwidth}{!}{
    \begin{tabular}{@{}l|c@{}}
\toprule
Notation & Description \\\midrule
$\mathcal{G}=(\mathcal{V}, \mathcal{E})$ & a graph composed of node set $\mathcal{V}$ and edge set $\mathcal{E}$ \\
$\mathbf{y}_v$; $\mathbf{y}_{\mathcal{G}}$ & labels of node $v$ and graph $\mathcal{G}$, respectively\\
$m_{\mathcal{N}(v)}$                     & message aggregated from $v$'s neighborhood $\mathcal{N}(v)$\\
$\mathbf{X}$; $\mathbf{H}^{(k)}$           & node embeddings of initial and $k$-th layers, respectively\\
$\mathbf{h}_v$; $\mathbf{z}_v$             & representation vector and logit of node $v$, respectively\\
$\mathbf{s}_\mathcal{G}$; $\mathbf{z}_\mathcal{G}$             & summary vector and logit of graph $\mathcal{G}$, respectively\\
$G^T$; $G^S$                               &  GNN models of teacher and student, respectively\\
$\mathbf{H}^T$; $\mathbf{H}^S$             & node embeddings of teacher and student, respectively\\
$\mathbf{Z}^T$; $\mathbf{Z}^S$             & logits of teacher and student, respectively\\
$D_e$; $D_\ell$ & identifiers of node embeddings and logits, respectively\\\bottomrule
    \end{tabular}
    }
\vspace{-3ex}
\end{table}

\section{Background}\label{sec:gnn}
In this part, we review the basic concepts of GNNs. Main notations are summarized in Table \ref{tab:notation}.
GNNs are designed as an extension of convolutions to non-Euclidean data \cite{DBLP:journals/corr/BrunaZSL13}.
In this paper, we mainly select message passing based GNNs \cite{DBLP:conf/icml/GilmerSRVD17} as both teacher and student models, {where} messages are exchanged between nodes and updated with neural networks \cite{DBLP:conf/icml/GilmerSRVD17}.
Let $\mathcal{G}=(\mathcal{V}, \mathcal{E})$ denote a graph with feature vector $\mathbf{X}_v$ for node $v\in \mathcal{V}$.
We are interested in two tasks, namely (1) \textit{Node classification}, where each node $v\in \mathcal{V}$ has an associated label $y_v$ and the goal is to learn a representation vector $\mathbf{h}_v$ of $v$ that aids in predicting $v$'s label as $\hat{\mathbf{y}}_v=f(\mathbf{h}_v)$; and (2) \textit{Graph classification}, where we are given a set of graphs $\{\mathcal{G}_1, \cdots, \mathcal{G}_N\}$ with corresponding labels $\{\mathbf{y}_{\mathcal{G}_1}, \cdots, \mathbf{y}_{\mathcal{G}_N}\}$ and the goal is to learn a summary vector $\mathbf{s}_\mathcal{G}$ such that the label of an entire graph can be predicted as $\hat{\mathbf{y}}_\mathcal{G}=f(\mathbf{s}_\mathcal{G})$.
Therefore, we decompose a general GNN into an encoder and a classifier.
The encoder follows a message-passing scheme, where the hidden embedding $\mathbf{h}_v^{(k+1)}$ of node $v\in \mathcal{V}$ is updated according to information aggregated from $v$'s graph neighborhood $\mathcal{N}(v)$ at $k$-th iteration.
This message-passing update \cite{grl} can be expressed as
\begin{align*}
    \mathbf{h}_v^{(k+1)} &=\textup{UPDATE}^{(k)}\left(\mathbf{h}_v^{(k)}, m_{\mathcal{N}(v)}^{(k)}\right), \\
    \textup{where } m_{\mathcal{N}(v)}^{(k)} &= \textup{AGGREGATE}^{(k)}\left( \{\mathbf{h}_u^{(k)}\mid \forall\,u\in \mathcal{N}(v) \} \right).
\end{align*}
Note that the initial embeddings are set to the input features for all the nodes if $k=0$, i.e., $\mathbf{h}_v^{(0)}=\mathbf{X}_v, \forall\,v\in \mathcal{V}$. 
After running $K$ iterations of the GNN message passing, we derive the final representation $\mathbf{h}_v=\mathbf{h}_v^{(K)}$ for each node $v\in \mathcal{V}$.
For graph-level classification, the READOUT function aggregates final node embeddings to obtain the summary vector $\mathbf{s}_\mathcal{G}$ of the entire graph, i.e., 
\[
\mathbf{s}_\mathcal{G}=\textup{READOUT}(\{\mathbf{h}_v\mid \forall\,v\in\mathcal{V}\}),
\]
where the READOUT function can be a simple permutation invariant function such as max-pooling and mean-pooling \cite{reddit, gin}.
The classifier then reads into the final representation of a node or a graph for node-level or graph-level classification, i.e.,
\begin{align*}
    \mathbf{z} &= g(\mathbf{h}), \\
    \hat{\mathbf{y}} &=\textup{softmax}(\mathbf{z}),
\end{align*}
where we usually interpret $\mathbf{z}_v$ (or $\mathbf{z}_\mathcal{G}$) and $\hat{\textbf{y}}_v$ (or $\hat{\textbf{y}}_\mathcal{G}$) as logit and prediction of a node (or a graph), respectively.

%% file: section/med.tex
\begin{figure*}[!t]
    \centering
    \includegraphics[width=0.85\linewidth]{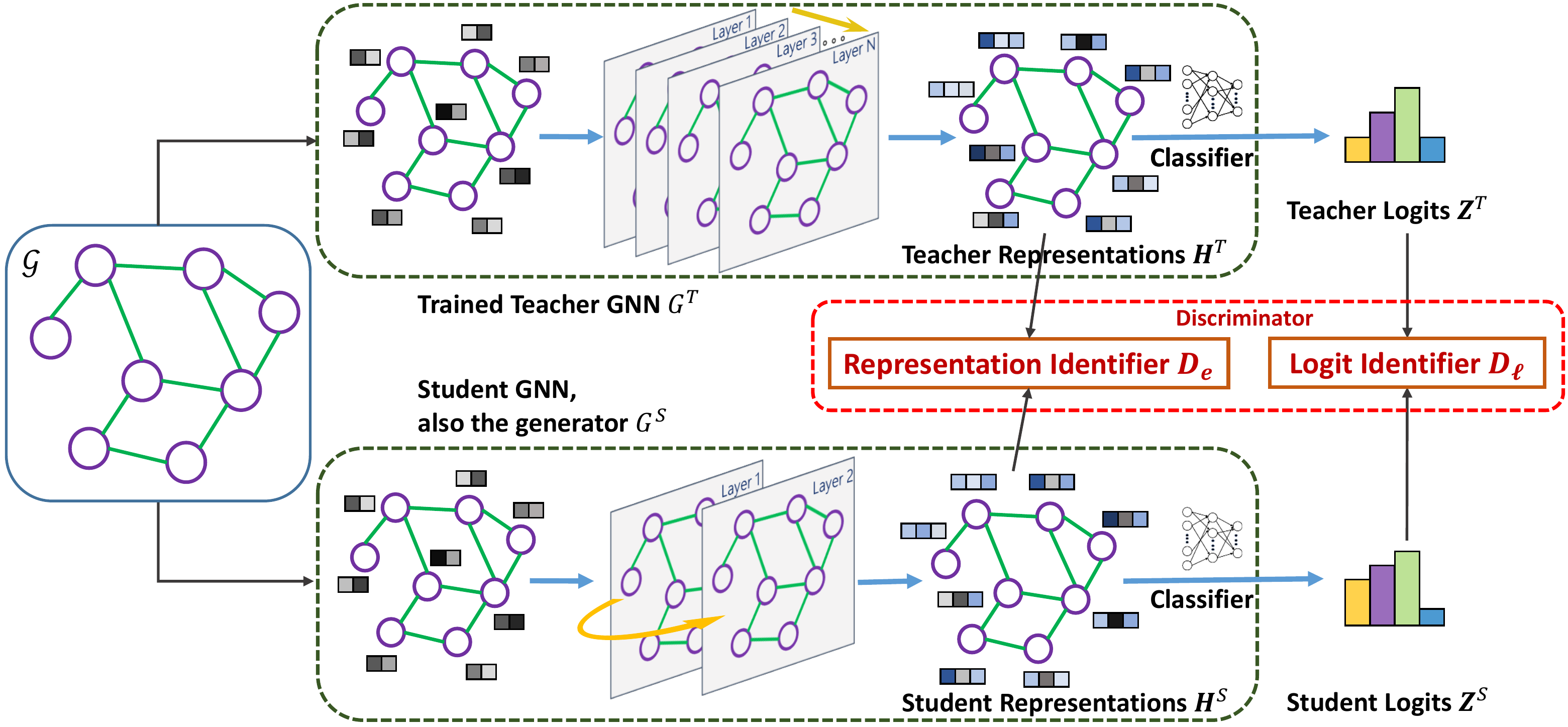}
    \vskip -0.5em
    \caption{Illustration of the proposed adversarial knowledge distillation framework \med.}
    \label{fig:framework}
    \vskip -0.5em
\end{figure*}

\section{Methodology}
In this part, we first introduce our adversarial knowledge distillation framework in Section \ref{sec:framework}. 
Next, we detail the proposed representation identifier in Section \ref{sec:rep-dis} and logit identifier in Section \ref{sec:log-dis}. 

\subsection{An Adversarial Knowledge  Distillation Framework for Graph Neural Networks} \label{sec:framework}

We note that a powerful knowledge distillation method in graph domains should be able to 
1) adaptively detect the difference between teacher and student on graphs of various fields; 2) inherit the teacher knowledge as much as possible; 
and 3) transfer teacher knowledge in a {topology-aware manner.}

Previous work \cite{DBLP:conf/icml/ChungPKK20,kdgan,DBLP:conf/iclr/0002HH18} has demonstrated that distance functions such as $\ell_p$ distance and KL-divergence are too vigorous for student models with a small capacity. Even worse, \citet{DBLP:conf/aaai/WangXXT18} declared that it is hard to determine which distance formulation is optimal.
Therefore, we develop the first adversarial distillation framework in graph domains to meet the first requirement.
%
Furthermore, inspired by the fact that intermediate representations can provide hints for knowledge transfer \cite{fitnets}, we take advantage of node representations and logits derived from deep teacher models to improve the training of compact student networks.
Finally, to meet the third requirement, we develop a topology-aware discriminator, which stimulates student networks to mimic teachers and produce similar local affinity of connected nodes and global affinity of patch-summary pairs.

Figure \ref{fig:framework} illustrates the overall architecture of \med, which adversarially trains the student model against a topology-aware discriminator in a two-player minimax game.
The student GNN serves as a generator and produces node embeddings and logits that are similar to teacher output, while the discriminator aims to distinguish between teacher output and what the generator produces.
The minimax game ensures the student network to perfectly model the probability distribution of teacher knowledge at the equilibrium via adversarial losses \cite{kdgan,DBLP:journals/pami/WangZSQ21}.

\begin{algorithm}[!ht]
    \caption{\med~for node-level classification.} \label{alg:framework}
    \begin{algorithmic}[1]
        \Require Graph $\mathcal{G}=(\mathcal{V}, \mathcal{E})$, adjacent matrix $\mathbf{A}$, node features $\mathbf{X}$ and the pretrained teacher model $G^T$.
        \Ensure The learnt student model $G^S$.
        \State $\mathbf{H}^T, \mathbf{Z}^T=G^T(\mathbf{X}, \mathbf{A})$; $\mathbf{s}^T=\frac{1}{|\mathcal{V}|}\sum_{v\in \mathcal{V}}\mathbf{h}_v^T$
        \While{not converge} 
            \State $\mathbf{H}^S, \mathbf{Z}^S=G^S(\mathbf{X}, \mathbf{A})$; $\mathbf{s}^S=\frac{1}{|\mathcal{V}|}\sum_{v\in \mathcal{V}}\mathbf{h}_v^S$
            \ForAll{node $v\in \mathcal{V}$}
                \State Update $D_e$ to distinguish $(\mathbf{h}_v^T, \mathbf{s}^T)$ and $(\mathbf{h}_v^S, \mathbf{s}^T)$
                \State Update $D_e$ to distinguish $(\mathbf{h}_v^S, \mathbf{s}^S)$ and $(\mathbf{h}_v^T, \mathbf{s}^S)$ 
                \For{node $u\in \mathcal{N}(v)$}
                    \State Update $D_e$ to distinguish $(\mathbf{h}_v^T, \mathbf{h}_u^T)$ and $(\mathbf{h}_v^S, \mathbf{h}_u^S)$
                    \State Update the parameters of $G^S$ to fool $D_e$ via Eq.~\ref{eq:emb_D}
                \EndFor
                \State Update $D_\ell$ to distinguish $\mathbf{z}_v^T$ and $\mathbf{z}_v^S$
                \State Update the parameters of $G^S$ to fool $D_\ell$ via Eq.~\ref{eq:logits_D}
            \EndFor
        \EndWhile 
        \State \Return $G^S$
    \end{algorithmic}
\end{algorithm}

In addition to teacher and student GNNs, \med~includes a novel discriminator as well, which can be decomposed into a representation identifier and a logit identifier.
As the node classification task is popular in existing distillation literature \cite{DBLP:conf/cvpr/YangQSTW20,DBLP:conf/sigmod/ZhangMSJCR020,DBLP:conf/www/0002LS21}, we take node classification for an example and represent the training procedure of \med~for node-level classification in
Algorithm \ref{alg:framework}.
As for the graph-level algorithm, please refer to Appendix \ref{sec:alg2}.
Let $D_e$ and $D_\ell$ be the two identifiers that operate on node embeddings and logits, respectively.
Suppose we have a pretrained teacher GNN $G^T$ and accompanying knowledge, i.e., node representations $\mathbf{H}^T$ and logits $\mathbf{Z}^T$.
The objective is to train a student GNN $G^S$ with much less parameters while preserving the expressive power of the teacher $G^T$.
That is, we expect that the student $G^S$ can generate high-quality node representations $\mathbf{H}^S$ and logits $\mathbf{Z}^S$ to achieve competitive performance against the teacher.
As the trainable discriminator is more flexible and tolerant than most specific distance including KL-divergence and $\ell_p$ distance \cite{DBLP:conf/icml/ChungPKK20,DBLP:conf/iclr/0002HH18}, \med~enables student GNNs to capture inter-node and inter-class correlations instead of mimicking the exact distribution of teacher knowledge.

\subsection{The Representation Identifier} \label{sec:rep-dis}
In this section, we introduce our topology-aware representation identifier.
Note that we decompose a GNN into an encoder and a classifier in Section \ref{sec:gnn}.
The representation identifier focuses on the output of the GNN encoder, i.e., the final node embeddings.
%
Instead of directly matching the feature maps of teacher and student \cite{DBLP:conf/cvpr/YangQSTW20}, we adversarially distill node representations of teacher models from both local and global views.

Given the node representations of teacher and student GNNs, namely $\mathbf{H}^T\in \mathbb{R}^{| \mathcal{V}|\times d^T}$ and $\mathbf{H}^S\in \mathbb{R}^{| \mathcal{V}|\times d^S}$, we perform mean-pooling to obtain the corresponding summary vectors for a graph as $\mathbf{s}^T\in \mathbb{R}^{d^T}$ and $\mathbf{s}^S\in \mathbb{R}^{d^S}$.
In practice we set the dimension of student's node representations equal to that of teacher's, i.e., $d^S=d^T=d$.
For each node $v$, the representation learned by the student GNN $G^S$ is criticized by the identifier $D_e$ through both local and global views.
Specifically, reading in connected node pair $\{\mathbf{h}_v, \mathbf{h}_u\}$ or patch-summary pair $\{\mathbf{h}_v, \mathbf{s}\}$, the identifier $D_e$ is expected to predict a binary value ``Real/Fake'' that indicates whether the pair is real or fake.
Fake node pair implies that the representations of the two adjacent nodes are produced by the student GNN $G^S$;
while fake patch-summary pair implies that the patch representation and summary representation are produced by GNNs of different roles.

Formally, the topology-aware identifier $D_e$ consists of $D_e^{local}$ and $D_e^{global}$. If nodes $v$ and $u$ form an edge on the graph $\mathcal{G}$, then $D_e^{local}$ maps representations of the two connected nodes to the real value that we interpret as affinity between connected nodes. On the other hand, for each node $v$ on the graph, $D_e^{global}$ maps the patch-summary pair to the real value that we interpret as affinity between node and graph. That is,
\begin{align*}
    D_e^{local}(\mathbf{h}_v^T, \mathbf{h}_u^T) &= \langle \mathbf{h}_v^T, \mathbf{W}^{local} \mathbf{h}_u^T \rangle \in [0, 1],\\
    D_e^{local}(\mathbf{h}_v^S, \mathbf{h}_u^S) &= \langle \mathbf{h}_v^S, \mathbf{W}^{local} \mathbf{h}_u^S \rangle \in [0, 1], \quad \forall~(v,u)\in \mathcal{E},\\
    D_e^{global}(\mathbf{h}_v^{T/S}, \mathbf{s}_\mathcal{G}^{T}) &= \langle \mathbf{h}_v^{T/S}, \mathbf{W}^{global} \mathbf{s}_\mathcal{G}^{T} \rangle \in [0, 1],\\
    D_e^{global}(\mathbf{h}_v^{T/S}, \mathbf{s}_\mathcal{G}^{S}) &= \langle \mathbf{h}_v^{T/S}, \mathbf{W}^{global} \mathbf{s}_\mathcal{G}^{S} \rangle \in [0, 1], \quad \forall~v\in \mathcal{V}\subset\mathcal{G},
\end{align*}
where $\textbf{W}^{local}$ and $\textbf{W}^{global}$ are learnable diagonal matrices.
By this means, $D_e^{local}$ encourages the student to inherit the local affinity hidden in teacher's node embeddings, while $D_e^{global}$ encourages the student to inherit the global affinity.

As the student GNN aims to fool the representation identifier $D_e$, we view $G^S$ as a generator. 
Under the guidance of the identifier $D_e$, the generator strives to yield indistinguishable node representations.
The adversarial training process can be formulated as a two-player minimax game, i.e.,
\begin{equation}
\begin{aligned}
    \min_{G^S} \max_{D_e} \mathcal{J}^{local} +\mathcal{J}^{global},
\end{aligned} 
\label{eq:emb_D}
\end{equation}
where $\mathcal{J}^{local}$ is written as
\begin{align*}
    \frac{1}{|\mathcal{E}|}\sum_{(v,u)\in \mathcal{E}}\Big( &\log\textup{P}(\textup{Real}\mid D_e^{l}(\mathbf{h}_v^T, \mathbf{h}_u^T))+\log\textup{P} (\textup{Fake}\mid D_e^{l}(\mathbf{h}_v^S, \mathbf{h}_u^S))\Big),
\end{align*}
and $\mathcal{J}^{global}$ is written as
\begin{align*}
    \frac{1}{2|\mathcal{V}|}\sum_{v\in \mathcal{V}}\Big( &\log\textup{P}(\textup{Real}\mid D_e^{g}(\mathbf{h}_v^T, \mathbf{s}_\mathcal{G}^T))+\log\textup{P} (\textup{Fake}\mid D_e^{g}(\mathbf{h}_v^S,\mathbf{s}_\mathcal{G}^T))+\\
    &\log\textup{P}(\textup{Real}\mid D_e^{g}(\mathbf{h}_v^S,\mathbf{s}_\mathcal{G}^S))
    +\log\textup{P} (\textup{Fake}\mid D_e^{g}(\mathbf{h}_v^T,\mathbf{s}_\mathcal{G}^S))\Big),
\end{align*}
where $D_e^{l}$ and $D_e^{g}$ denote $D_e^{local}$ and $D_e^{global}$, respectively.
By alternately maximizing and minimizing the objective function, we finally obtain an expressive student GNN when it converges.

It is worth noting that we can understand the representation identifier $D_e$ from different perspectives.
In fact, $D_e^{local}$ can degenerate into a bilinear distance function. If we modify the trainable diagonal matrix to the identity matrix and normalize the input vectors, 
then the local affinity calculated by $D_e^{local}$ is equivalent to cosine similarity between node embeddings. That is, if $\textbf{W}^{local}=\textbf{I}$ and $\widehat{\mathbf{h}}_v=\mathbf{h}_v/\|\mathbf{h}_v\|_2$, then
\begin{align*}
    \langle \widehat{\mathbf{h}}_v, \textbf{W}^{local}  \widehat{\mathbf{h}}_u \rangle=\textup{cosine\_sim}(\mathbf{h}_v, \mathbf{h}_u), ~\forall~(v,u)\in \mathcal{E}.
\end{align*}
Apart from that, we can perceive the training process of $D_e^{global}$ as the maximization of the mutual information between the graph-level representation (i.e., the summary vector $\mathbf{s}_\mathcal{G}$) and the node-level representation (i.e., the patch vector $\mathbf{h}_v$).
Likewise, \citet{dgi} propose a discriminator to maximize the mutual information between graph representations of different levels. However, our proposed $D_e^{global}$ differs from DGI \cite{dgi} in two aspects.
First, the performance of DGI \cite{dgi} heavily relies on how to draw negative samples, while $D_e^{global}$ exempts the need of negative sampling.
Second, DGI \cite{dgi} judges the representations produced by the same encoder for unsupervised graph learning, while $D_e^{global}$ discriminates the representations produced by teacher and student for adversarial knowledge distillation.

Compared with existing distance-based embedding distillation in graph domains \cite{DBLP:conf/cvpr/YangQSTW20}, we replace predefined distance formulations with a more flexible and tolerant identifier.
Instead of mimicking the exact distribution of teacher node embeddings, our representation identifier enables student GNN to capture the inter-node correlation, which is proved to be of crucial importance in graph domains \cite{DBLP:conf/wsdm/JinDW0LT21}.

\subsection{The Logit Identifier} \label{sec:log-dis}
In this section, we introduce the second identifier of the proposed framework, which operates on the output logits.
For notation convenience, we introduce the logit identifier in the context of node-level classification.
Output of the GNN-based classifier is a probability distribution over categories. The probability is derived by applying a softmax function over the output of the last fully connected layer, which is also known as logits.
By leveraging adversarial training, we aim to transfer inter-class correlation \cite{DBLP:conf/aaai/YangXQY19} from complicated teacher GNNs to compact student GNNs.

Instead of forcing the student to exactly mimic the teacher by minimizing KL-divergence \cite{hinton2015distilling} or other predefined distance, we transfer the inter-class correlation hidden in teacher logits through a logit identifier.
Inspired by adversarial training in visual representation learning \cite{DBLP:conf/aaai/WangXXT18,DBLP:conf/iclr/0002HH18}, our logit identifier is trained to distinguish student logits from teacher logits, while the generator (i.e., the student GNN) is adversarially trained to fool the identifier. That is, we expect the compact student GNN to output logits similar to the teacher logits so that the identifier cannot distinguish.

As residual learning can mitigate the gap between teacher and student \cite{DBLP:journals/ijon/GaoWW21},
we use an MLP with residual connections as our logit identifier $D_\ell$.
The number of hidden units in each layer is the same as the dimension of logit, which is equal to the number of categories $C$.
A plain identifier reads into the logit of each node and predicts a binary value ``Real/Fake'' that indicates whether the logit is derived by teacher or student.
Denote the logit of node $v$ derived by teacher and student as $\mathbf{z}_v^T$ and $\mathbf{z}_v^S$, respectively.
A plain identifier $D$ aims to maximize the log-likelihood as
\begin{equation}
\max_D \frac{1}{| \mathcal{V}|}\sum_{v\in \mathcal{V}}
\Big(\log \textup{P}(\textup{Real}\mid D(\mathbf{z}_v^T))+\log \textup{P}(\textup{Fake}\mid D(\mathbf{z}_v^S))\Big).
\end{equation}
As \citet{DBLP:conf/iclr/0002HH18} pointed out that the plain version is slow and unstable, we follow \cite{DBLP:conf/icml/OdenaOS17,DBLP:conf/iclr/0002HH18} and modify the objective to also predict the specific node labels.
Therefore, the output of $D_\ell$ is a $C+1$ dimensional vector with the first $C$ for label prediction and the last for Real/Fake (namely teacher/student) indicator.
We thus maximize following objective for the training of $D_\ell$:
\begin{equation}
\begin{aligned}
\max_{D_\ell} \frac{1}{| \mathcal{V}|}\sum_{v\in \mathcal{V}}&
\Big(\log \textup{P}(\textup{Real}\mid D_\ell(\mathbf{z}_v^T))+\log \textup{P}(\textup{Fake}\mid D_\ell(\mathbf{z}_v^S))\\
&+\log \textup{P}(y_v\mid D_\ell(\mathbf{z}_v^T))+\log \textup{P}(y_v\mid D_\ell(\mathbf{z}_v^S))\Big).    
\end{aligned}
\label{eq:logits-d}
\end{equation}

As for the training of generator, i.e., the student GNN $G^S$, we follow \cite{DBLP:conf/cvpr/IsolaZZE17,DBLP:conf/iclr/0002HH18} and introduce instance-level alignment between teacher and student logits besides the category-level alignment. 
We thus minimize the loss function for the training of $G^S$ as
\begin{equation}
\begin{aligned}
\min_{G^S} \frac{1}{| \mathcal{V}|}\sum_{v\in \mathcal{V}}&
\Big(\log \textup{P}(\textup{Real}\mid D_\ell(\mathbf{z}_v^T))+\log \textup{P}(\textup{Fake}\mid D_\ell(\mathbf{z}_v^S))\\
&-\Big[ \log \textup{P}(y_v\mid D_\ell(\mathbf{z}_v^T))+\log \textup{P}(y_v\mid D_\ell(\mathbf{z}_v^S)) \Big]\\
&+\|\mathbf{z}_v^S - \mathbf{z}_v^T\|_1
\Big).  
\end{aligned} 
\label{eq:logits_D}
\end{equation}

Compared to existing knowledge distillation models for GNNs \cite{DBLP:conf/sigmod/ZhangMSJCR020,DBLP:conf/www/0002LS21}, the logit identifier relaxes the rigid coupling between student and teacher.
Besides, the adversarial training approach relieves the pain for hand-engineering the loss.


%% file: section/exp.tex
\section{Experiments}

In this part, we conduct extensive experiments to evaluate the capability of our proposed \med. Our experiments are intended to answer the following five research questions.

\vspace{2mm}
\noindent \textbf{RQ1:} How does \med~perform on node-level classification?

\noindent \textbf{RQ2:} How does \med~perform on graph-level classification?

\noindent \textbf{RQ3:} How efficient are the student GNNs trained by \med?

\noindent \textbf{RQ4:} How do different components (i.e., $D_e$ or $D_\ell$) affect the performance of \med? 

\noindent \textbf{RQ5:} Do student GNNs learn better node representations when equipped with \med? 

\begin{table}[!t]\small
\caption{Statistics of the eight node classification benchmarks.}
\centering
\begin{tabular}{@{}lcccc@{}}
\toprule
\textbf{Datasets} & \textbf{\#Nodes} & \textbf{\#Edges} & \textbf{\#Feat.} & \textbf{Data Split} \\\midrule
Cora \cite{DBLP:journals/ir/McCallumNRS00,DBLP:conf/iclr/BojchevskiG18} & 2,708            & 5,429            & 1,433      & 140/500/1K \\
CiteSeer \cite{DBLP:journals/aim/SenNBGGE08}& 3,327            & 4,732            & 3,703           & 120/500/1K \\
PubMed \cite{namata2012query}& 19,717           & 44,338           & 500    & 60/500/1K \\
\midrule
Flickr \cite{DBLP:conf/eccv/McAuleyL12,graphsaint}& 89,250           & 899,756          & 500   & 44K/22K/22K \\
Arxiv \cite{ogb}& 169,343          & 1,166,243        & 128 & 90K/29K/48K \\
Reddit \cite{reddit,graphsaint}& 232,965          & 23,213,838       & 602  & 153K/23K/55K \\
\midrule
Yelp \cite{graphsaint}& 716,847          & 13,954,819       & 300     & 537K/107K/71K \\
Products \cite{ogb}& 2,449,029        & 61,859,140       & 100     & 196K/39K/2M \\
\bottomrule
\end{tabular}
\vspace{-3ex}
\label{tab:dataset}
\end{table}

\subsection{Experimental Setup} \label{sec:setup}

\begin{table*}[!ht]\small
\caption{Performance on Node Classification (metric: F1-micro (\%) ). 
``O. Perf.'' and ``R. Perf.'' refer to performance reported in original papers and reproduced by our own, respectively. Higher of these two columns are underlined.
``Perf. Impv.'' and ``\#Params Decr.'' refer to the absolute improvement of student performance (w.r.t. the underlined results) and the relative decrease of teacher parameters, respectively.
Results of previous work are mainly taken from \cite{graphsaint}, \cite{gas}, and OGB Leaderboards. We report the average performance and std. across 10 random seeds.}
\centering
\begin{tabular}{@{}l  ccc ccc cccc@{}}
\toprule
&  \multicolumn{3}{c}{\textbf{Teacher}} & \multicolumn{3}{c}{\textbf{Vanilla Student}} & \multicolumn{4}{c}{\textbf{Student trained with \med}} \\ 
\cmidrule(lr){2-4}
\cmidrule(lr){5-7}
\cmidrule(lr){8-11}
Datasets& Model & Perf.& \#Params  & Model&O. Perf. & R. Perf. & Perf.      & \#Params  & Perf. Impv. (\%) & \#Params Decr. \\\midrule
Cora    & GCNII & 85.5 & 616,519   & GCN  & \underline{81.5}    & 78.3 $\pm$0.9    & 83.6 $\pm$0.8   & 96,633    & 2.1   & 84.3\% \\
CiteSeer& GCNII & 73.4 & 5,144,070 & GCN  & \underline{71.1}    & 68.6 $\pm$1.1    & 72.9 $\pm$0.4   & 1,016.156 & 1.8   & 80.2\% \\
PubMed  & GCNII & 80.3 & 1,177,603 & GCN  & \underline{79.0}    & 78.1 $\pm$1.0    & 81.3 $\pm$0.4   & 195,357   & 2.3   & 83.4\% \\ \midrule
Flickr  & GCNII & 56.20& 1,182,727 & GCN  & 49.20   & \underline{49.63} $\pm$1.19  & 52.95 $\pm$0.24 & 196,473   & 3.32  & 83.4\% \\
Arxiv   & GCNII & 72.74& 2,148,648 & GCN  & \underline{71.74}   & 71.43 $\pm$0.13  & 73.05 $\pm$0.22 & 242,426   & 1.31  & 88.7\% \\
Reddit  & GCNII & 96.77& 691,241   & GCN  & 93.30   & \underline{94.12} $\pm$0.04  & 95.15 $\pm$0.02 & 234,655   & 1.03  & 66.1\% \\ \midrule
Yelp    & GCNII & 65.14& 2,306,660 & Cluster-GCN& 59.15 & \underline{59.63} $\pm$0.51 & 60.63 $\pm$0.42 & 431,950   & 1.00  & 81.3\% \\
Products& GAMLP & 84.59& 3,335,831 & Cluster-GCN& \underline{76.21} & 74.99 $\pm$0.76 & 81.45 $\pm$0.47 & 682,449   & 5.24  & 79.5\% \\
\bottomrule
\end{tabular}
\vspace{-1ex}
\label{tab:results}
\end{table*}

\subsubsection{Datasets.}
For a comprehensive comparison, in Section \ref{sec:node-level results} we perform node classification on eight widely-used datasets, covering graphs of various sizes. 
The statistics are summarized in Table \ref{tab:dataset}.
To evaluate the effectiveness of \med~on graph-level classification, we benchmark \med~against traditional knowledge distillation methods on two molecular property prediction datasets \cite{ogb} in Section \ref{sec:graph-level results}.
For detailed information on the ten datasets, please refer to Appendix \ref{app:dataset}.
In a nutshell, all datasets are collected from real-world networks in different domains, including social networks, citation networks, molecular graphs, and trading networks. We conduct experiments under both transductive and inductive settings, involving both textual and visual features.

\subsubsection{Model Selection for Student and Teacher.}
In fact, \med~is applicable to all message passing based GNNs.
For node-level classification, we select two simple and famous GNNs as the student models. 
Specifically, we choose GCN \cite{gcn} for Cora \cite{DBLP:journals/ir/McCallumNRS00,DBLP:conf/iclr/BojchevskiG18}, CiteSeer \cite{DBLP:journals/aim/SenNBGGE08}, PubMed \cite{namata2012query}, Flickr \cite{DBLP:conf/eccv/McAuleyL12,graphsaint}, Arxiv \cite{ogb} and Reddit \cite{reddit,graphsaint}, while Cluster-GCN \cite{cluster-gcn} is selected for large-scale datasets including Yelp \cite{graphsaint} and Products \cite{ogb}.
On the other hand, we employ two deep teacher GNNs on different graphs, namely GAMLP \cite{gamlp} for Products and GCNII \cite{gcn2} for other seven datasets. 
As for graph-level classification, we test both GCN \cite{gcn} and GIN \cite{gin} as students on Molhiv \cite{ogb} and Molpcba \cite{ogb}. Simultaneously, we choose HIG\footnote{\url{https://github.com/TencentYoutuResearch/HIG-GraphClassification}.} as the sole teacher model for the graph-level task. Note that we execute the teacher models during pre-computation, which prepares node representations and logits for student training. For more implementation details and the information on the mentioned teacher and student graph models, please refer to Appendix \ref{app:detail}.

\subsection{Performance on Node Classification (RQ1)} \label{sec:node-level results} 
To evaluate the capability of our proposed adversarial knowledge distillation framework, we conduct node classification across graphs of various sizes.
Empirical results on eight datasets are presented in Table \ref{tab:results}.
Note that we follow the standard data split of previous work \cite{gcn,cluster-gcn,graphsaint,gas}.
As a node in the graph may belong to multiple classes (e.g., Yelp \cite{graphsaint}), we use F1-micro score to measure the performance.

In general, deep and complex GNNs have great expressive power and perform well on node classification task, especially for large graphs.
For example, GCNII \cite{gcn2}  outperforms the vanilla GCN \cite{gcn} by a large margin.
However, deep and wide GNNs usually suffer from prohibitive time and space complexity \cite{revgnn}.
By contrast, shallow and thin GNNs have small model capacity while they can easily scale to large datasets.
Tabel \ref{tab:results} shows that the proposed \med~enables shallow student GNNs to achieve comparable or even superior performance to their teachers while maintaining the computational efficiency.
Concretely, the GCN student outperforms the over-stacked teacher (i.e., GCNII) on PubMed and Arxiv with the knowledge transferred by \med.
As for large-scale graphs such as Flickr and Products, \med~consistently and significantly improves the performance of student GNNs.
We notice that to achieve superior accuracy on PubMed and Arxiv, the student GCN only requires less than 20\% parameters of its teacher.
However, the reason why \med~can achieve superior performance to the teacher GNNs is intriguing.
\citet{cheng2020explaining} have pointed out that knowledge distillation makes students learn various concepts simultaneously, rather than learn concepts from raw data sequentially. 
Therefore, \med~enables students learn both node-level and class-level views simultaneously, alleviating the problem that deep teacher GNNs tend to gradually discard views through layers according to the information-bottleneck theory \cite{wolchover2017new,shwartz2017opening}.

To further demonstrate the effectiveness of \med, we compare the proposed framework with several distillation methods including the traditional logit-based knowledge distillation (KD) \cite{hinton2015distilling}, the feature mimicking algorithm FitNet \cite{fitnets} and the recent local structure preserving (LSP) method \cite{DBLP:conf/cvpr/YangQSTW20}.
We reproduce them and present comparisons in Table \ref{tab:comparison}. Highest performance of each row is highlighted with boldface.
As graphs in different fields own distinct feature spaces, the three distance-based approaches fail to yield consistent improvement across all fields, which support the claim that a predefined and fixed distance is unfit to measure the discrepancy between teacher and student in different feature spaces.
Specifically, KD \cite{hinton2015distilling} performs even inferior than the vanilla student on Yelp; LSP \cite{DBLP:conf/cvpr/YangQSTW20} achieves poor performance on two trading networks (namely Yelp and Products); and FitNet \cite{fitnets} barely yields gains on two citation networks (namely CiteSeer and Arxiv).
On the other hand, \med~outperforms three baselines by a large margin on most of the eight datasets. 
Moreover, we note that LSP incurs the out-of-memory (OOM) issue on Arxiv and Reddit, while the GCN students equipped with \med~survive.

\begin{table}[!t]\small
\caption{Comparison with other distillation algorithms.}
\vspace{-1ex}
\centering
\begin{tabular}{@{}lccccc@{}}
\toprule
Datasets & Student & KD \cite{hinton2015distilling} & FitNet \cite{fitnets} & LSP \cite{DBLP:conf/cvpr/YangQSTW20} & \med  \\
\midrule
Cora     & 81.5  & 83.2  & 82.4  & 81.7   & \textbf{83.6} \\
CiteSeer & 71.1  & 71.4  & 71.6  & 68.8   & \textbf{72.9} \\
PubMed   & 79.0  & 80.3  & \textbf{81.3}  & 80.8   & \textbf{81.3} \\\midrule
Flickr   & 49.20  & 50.58 & 50.69 & 50.02  & \textbf{52.95} \\
Arxiv    & 71.74 & 73.03 & 71.83  & OOM    & \textbf{73.05} \\
Reddit   & 93.30  & 94.01 & 94.99  & OOM    & \textbf{95.15} \\\midrule 
Yelp     & 59.15 & 59.14 & 59.92  & 49.24  & \textbf{60.63} \\
Products & 76.21 & 79.19 & 76.57  & 70.86  & \textbf{81.45} \\
\bottomrule
\end{tabular}
\label{tab:comparison}
\end{table}

To understand why the proposed framework outperforms other baselines, we delve deeper into the four competitors.
We note that KD \cite{hinton2015distilling}, FitNet \cite{fitnets} and LSP \cite{DBLP:conf/cvpr/YangQSTW20} take advantage of teacher knowledge from different aspects.
Specifically, KD \cite{hinton2015distilling} only uses teacher logits as soft targets, while FitNet \cite{fitnets} and LSP \cite{DBLP:conf/cvpr/YangQSTW20} are proposed to leverage intermediate node embeddings.
Contrary to them, \med~leverages both aspects of teacher knowledge to distill inter-class and inter-node correlations.
Another important reason comes from the different ways they transfer teacher knowledge to student.
Concretely, KD \cite{hinton2015distilling}, FitNet \cite{fitnets} and LSP \cite{DBLP:conf/cvpr/YangQSTW20} force the student to mimic the exact distribution of teacher output (logits or intermediate embeddings) with fixed distance formulations, namely KL-divergence, Euclidean distance, and kernel functions.
\med~differs from KD \cite{hinton2015distilling}, FitNet \cite{fitnets} and LSP \cite{DBLP:conf/cvpr/YangQSTW20} as it transfers teacher knowledge via adversarial training, which is more tolerant and is less sensitive to parameters than the metric selection or temperature setting in traditional distance-based distillation.

\begin{table}[!ht]
    \centering
    \caption{Graph classification on Molhiv \cite{ogb} (metric: ROC-AUC (\%)) and Molpcba \cite{ogb} (metric: AP (\%)). Results of teacher and student are taken from OGB Leaderboards.
    We report the average performance and std. across 10 random seeds.
    }
    \resizebox{\linewidth}{!}{
    \begin{tabular}{@{}lcccc@{}}
        \toprule
        \textbf{Dataset} & \multicolumn{2}{c}{\textbf{Molhiv}} & \multicolumn{2}{c}{\textbf{Molpcba} } \\
        \textbf{Teacher} & \multicolumn{2}{c}{\textbf{HIG with DeeperGCN } } & \multicolumn{2}{c}{\textbf{HIG with Graphormer } } \\
        \textbf{Student} & \textbf{GCN} & \textbf{GIN} & \textbf{GCN} & \textbf{GIN} \\
        \midrule \midrule
        Teacher & 84.03 \text{$\pm$0.21} & 84.03 \text{$\pm$0.21} & 31.67 \text{$\pm$0.34} & 31.67 \text{$\pm$0.34} \\
        Student & 76.06 \text{$\pm$0.97} & 75.58 \text{$\pm$1.40} & 20.20 \text{$\pm$0.24} & 22.66 \text{$\pm$0.28} \\ \midrule
        KD \cite{hinton2015distilling} & 74.98 \text{$\pm$1.09} & 75.08 \text{$\pm$1.76} & 21.35 \text{$\pm$0.42} & 23.56 \text{$\pm$0.16} \\
        FitNet \cite{fitnets} & 79.05 \text{$\pm$0.96} & 77.93 \text{$\pm$0.61} & 21.25 \text{$\pm$0.91} & 23.74 \text{$\pm$0.19}\\
        \med & \textbf{79.46} \text{$\pm$0.97}& \textbf{79.16} \text{$\pm$1.50} & \textbf{22.56} \text{$\pm$0.23}& \textbf{25.85} \text{$\pm$0.17}\\
        \bottomrule
    \end{tabular}
    }
    \label{tab:molecules}
\end{table}

\subsection{Performance on Graph Classification (RQ2)}\label{sec:graph-level results}
To evaluate the capability of \med~on graph-level tasks, we conduct molecular property prediction across two benchmarks, namely Molhiv \cite{ogb} and Molpcba \cite{ogb}.
Table \ref{tab:molecules} summarizes the empirical results.
Note that the teacher model HIG on Molhiv \cite{ogb} is based on DeeperGCN \cite{li2020deepergcn}, while the teacher model on Molpcba \cite{ogb} is built upon Graphormer \cite{ying2021transformers}.
Our student architectures are GCN and GIN, which do not use virtual nodes and are considered to be less expressive but more efficient than HIG.

In Table \ref{tab:molecules}, we compare the proposed \med~with knowledge distillation approaches including the traditional logit-based KD \cite{hinton2015distilling} and the representation-based FitNet \cite{fitnets}.
We observe that \med~consistently improves student performance and outperforms the two distillation baselines.
Notably, as Graphormer \cite{ying2021transformers} is different from typical GNNs, both KD and FitNet yield minor performance boosts on Molpcba \cite{ogb}.
However, \med~yields significant improvement against the vanilla student models across the two graph-level benchmarks, which implies that \med~is promising to bridge graph models of different architectures.

\begin{figure*}[!ht]%
     \centering
     \subfloat[Vanilla GCN]{{\includegraphics[width=0.33\linewidth]{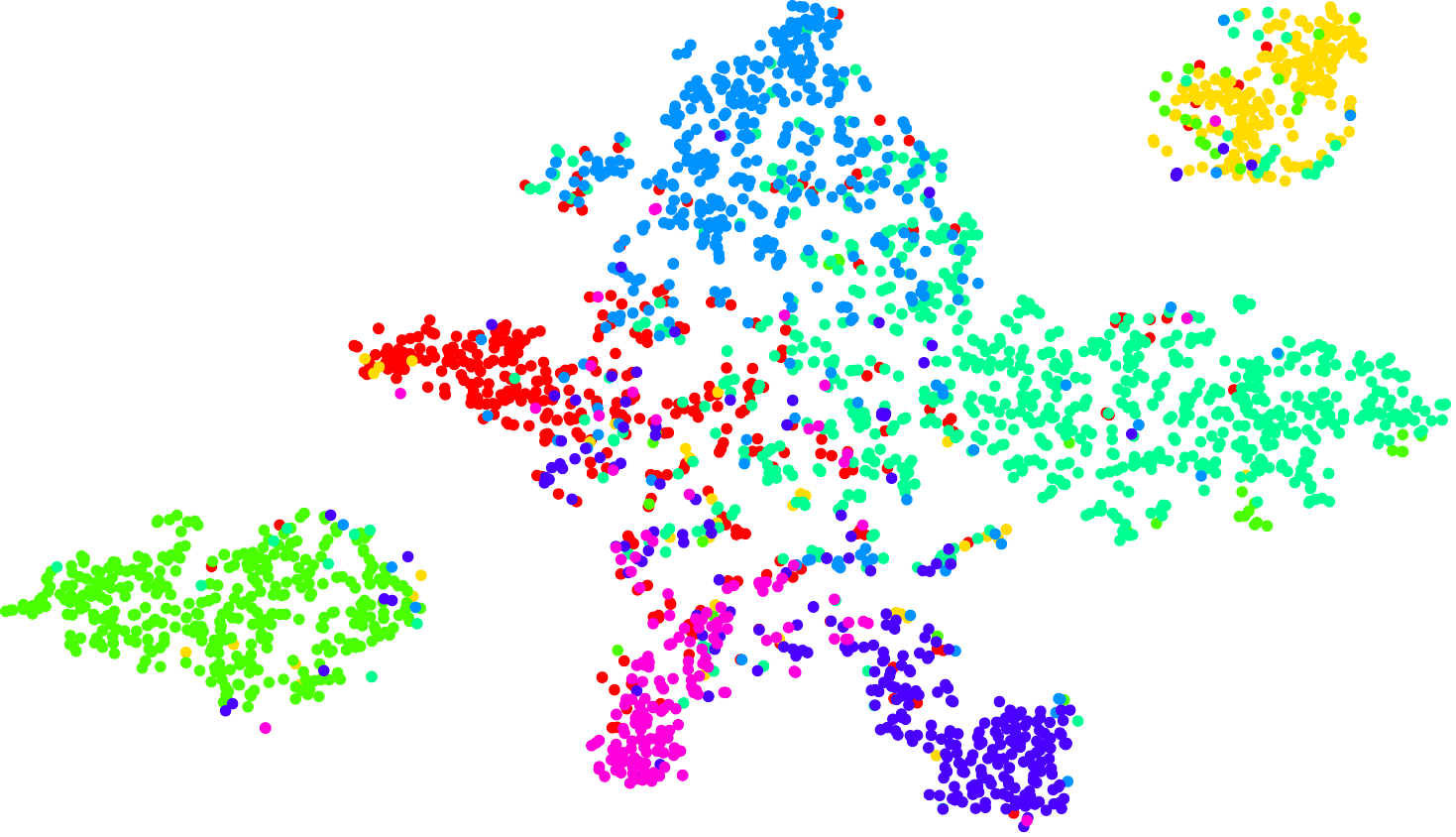} }\label{fig:gcn}}%
     \subfloat[Student GCN trained with \med]{{\includegraphics[width=0.33\linewidth]{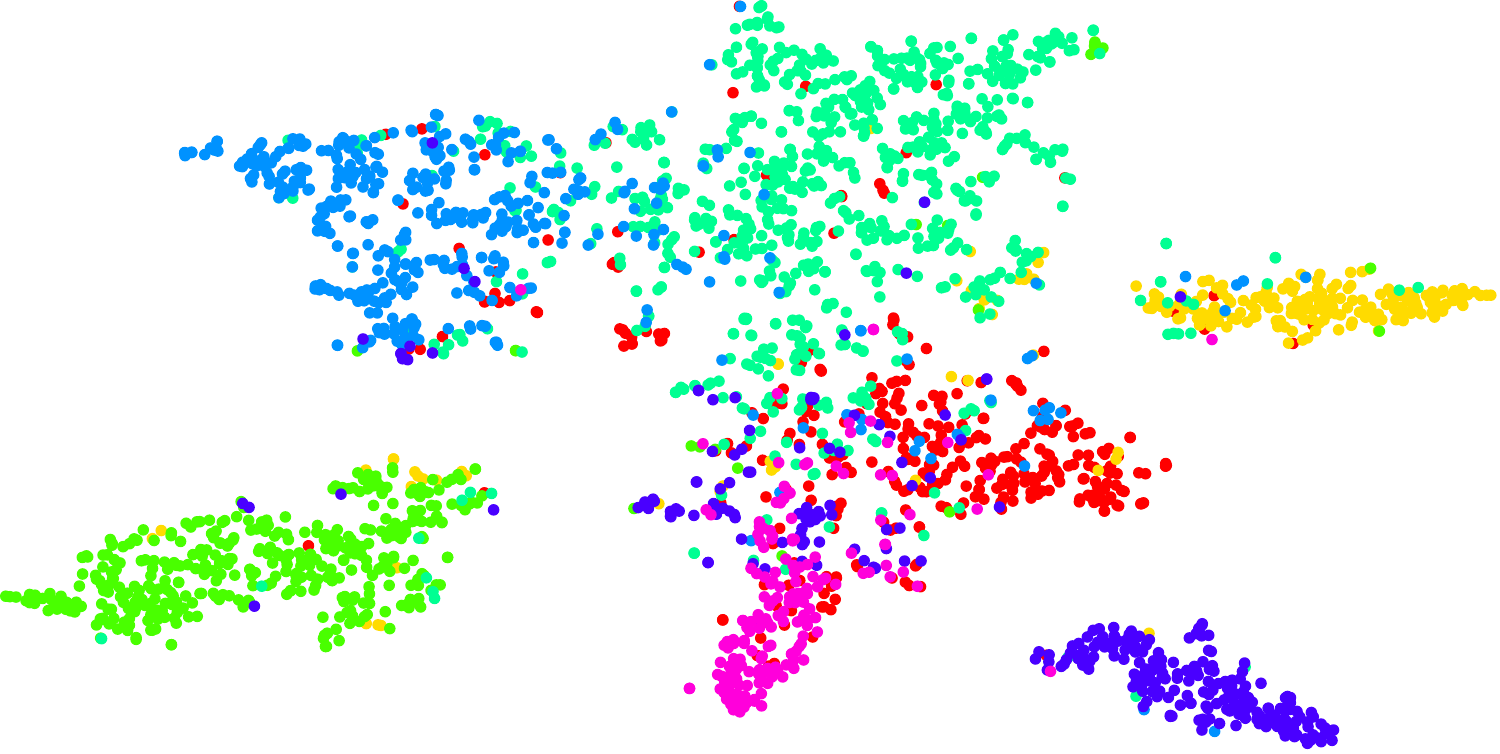} }\label{fig:stu}}%
    \subfloat[Teacher GCNII]{{\includegraphics[width=0.33\linewidth]{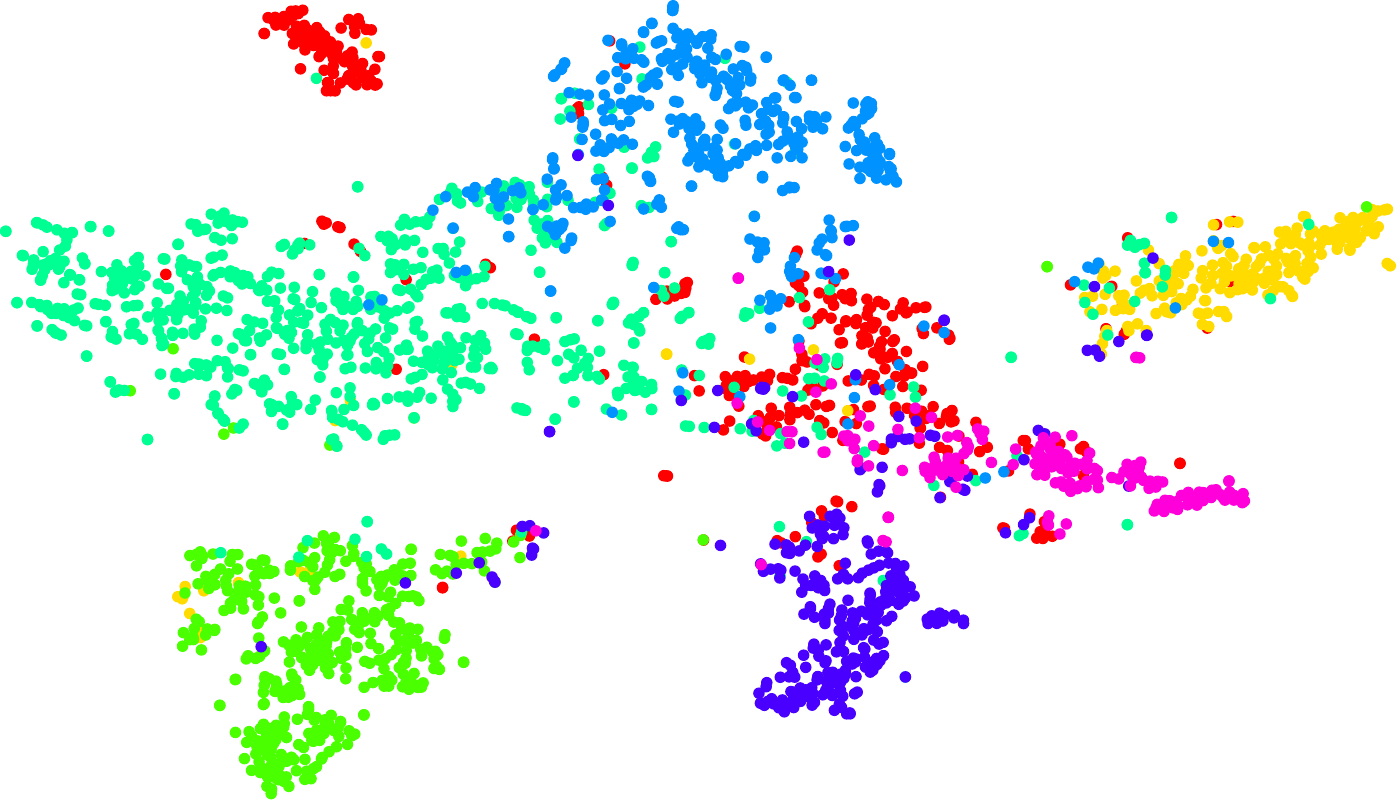} }\label{fig:tea}}%
    \qquad
    \vskip -0.5em
    \caption{t-SNE embeddings of the nodes in the Cora dataset from the vanilla GCN embeddings ({\bf left}), embeddings from the student GCN that trained by \med~({\bf middle}), and GCNII ({\bf right}). The Silhouette scores \cite{rousseeuw1987silhouettes} of the embeddings learned by three models are 0.2196, 0.2638, and 0.3033, respectively.}
    \label{fig:tsne}
\vskip -0.75em
\end{figure*}

\subsection{Analysis of Model Efficiency (RQ3)}
\label{sec:efficiency}
For practical applications, apart from effectiveness, the usability of a neural network also depends on model efficiency.
We investigate the efficiency of \med~with three criteria: 1) parameter-efficiency, 2) computation-efficiency, and 3) time-efficiency.
Specifically, we compare the student GNNs trained by \med~against their teachers in terms of the aforementioned three criteria.
We select five graph benchmarks (namely Cora, PubMed, Flickr, Yelp, and Products) and
summarize the efficiency comparison in Table \ref{tab:eff}.
Note that we conduct full-batch testing for both teacher and student on Cora, PubMed and Flickr. As for Yelp and Products, we align the batch size of teacher and student for a fair comparison.
We provide detailed analyses as follows.

\subsubsection{Parameter efficiency}\label{sec:param} For resource limited device, the amount of memory occupied by the model becomes critical to deployment. Here we use the number of parameters to measure the memory consumption for both teacher and student models. We observe that the number of node features, network layers, and hidden dimensions contribute to the number of parameters. By drastically lessening hidden dimensions and network layers, \med~reduces the model size of teacher GNNs to less than 20\% on average.

\begin{table}[!t]\small
\caption{Comparison of efficiency between student GNNs and the corresponding teachers.}
\centering
\resizebox{\columnwidth}{!}{
\begin{tabular}{@{}lcccccc@{}}
\toprule
& \multicolumn{2}{c}{\textbf{\#Params}} & \multicolumn{2}{c}{\textbf{GPU Memory}} &\multicolumn{2}{c}{\textbf{Inference time}} \\ 
\cmidrule(lr){2-3}
\cmidrule(lr){4-5}
\cmidrule(lr){6-7}
Datasets & Teacher & Student & Teacher & Student & Teacher & Student  \\
\midrule
Cora           & 0.6M & 0.1M & 0.22G & 0.03G & 40.3ms  & 4.1ms \\
PubMed         & 1.2M & 0.2M & 1.23G & 0.33G & 57.3ms  & 5.7ms \\
Flickr         & 1.2M & 0.2M & 2.79G & 1.49G & 309.7ms & 11.9ms\\
Yelp           & 2.3M & 0.4M & 6.28G & 4.73G & 3.0s & 1.5s \\
Products  & 3.3M & 0.7M & 6.25G & 6.20G & 16.1s   & 7.0s \\
\bottomrule
\end{tabular}
}
\vspace{-4ex}
\label{tab:eff}
\end{table}

\subsubsection{Computation efficiency}In addition to storage consumption, the dynamic memory usage is important as well. The reported GPU memory is the peak GPU memory usage during the first training epoch, which is also used in \cite{revgnn}. Restricted to the size and weight of mobile or embedded systems, neural networks that consume huge memory may be inapplicable to many practical situations.
We note that the GPU memory is highly related to the hidden dimensions and the number of network layers. As mentioned in Section \ref{sec:param}, \med~significantly reduces hidden dimensions and network layers of over-stacked teacher GNNs, which brings about portable and flexible student graph models.
\subsubsection{Time efficiency}As low latency is sometimes demanded in real-world applications, we also evaluate the inference time of both teacher and student GNNs, which refers to the time consumption of inference on testing dataset. 
Table \ref{tab:eff} shows that \med~significantly reduces the inference time for deep GNNs. On Flickr, in particular, \med~even cuts down more than 90\% inference time of the GCNII teacher, which shows that \med~trades off a reasonable amount of accuracy to reduce latency. The acceleration may originate from the sharp decrease in the number of network layers.
In a nutshell, Table \ref{tab:eff} shows that \med~is able to compress over-parameterized teacher GNNs into compact and computationally efficient student GNNs with relatively low latency.

\begin{table}[!t]\small
\caption{Ablation studies on the impacts of each identifier. }
\vspace{-1ex}
\centering
\begin{tabular}{@{}lcccccc@{}}
\toprule
Datasets & Cora  & PubMed & Flickr & Yelp  & Products & Molhiv \\\midrule
Teacher  & 85.5 & 80.3 & 56.20 & 65.14 & 84.59 & 84.03\\
Student  & 81.5 & 79.0 & 49.20 & 59.15 & 76.21 & 75.58\\\midrule
Only $D_e$    & 82.9 & 80.6 & 52.20 & 59.63 & 81.13 & 78.28\\
Only $D_\ell$ & 82.3 & 81.0 & 52.52 & 60.03 & 79.76 & 78.09\\
\med & \textbf{83.6} &\textbf{81.3} & \textbf{52.95} & \textbf{60.63} & \textbf{81.45} & \textbf{79.16} \\
\bottomrule
\end{tabular}
\vspace{-2ex}
\label{tab:abalation}
\end{table}

\subsection{Ablation Studies (RQ4)} \label{sec:ablation}
To thoroughly evaluate our framework, in this section we provide ablation studies to show the influence of the two identifiers.
Specifically, we separately test the capability of the two identifiers ($D_e$ and $D_\ell$) to clarify the essential improvement of each component.
We conduct our analysis on the same five node-level benchmarks used in Section \ref{sec:efficiency} as well as a graph-level dataset.
Highest performance of each column is highlighted in Table \ref{tab:abalation}.

We can conclude that the improvements benefit from both the representation identifier and the logit identifier.
Another valuable observation in Table \ref{tab:abalation} is that both $D_e$ and $D_\ell$ enable a vanilla GCN to achieve results superior to the GCNII teacher on PubMed.
The fact that \med~outperforms each of the identifiers demonstrates the importance of both node representations and logits. Either of the two identifiers captures orthogonal yet valuable knowledge via adversarial training. Specifically, the representation identifier excels at capturing inter-node correlation, while the logit identifier specializes in capturing inter-class correlation.

\subsection{Visualization (RQ5)} \label{sec:visualization}
We further perform qualitative analysis on the embeddings learnt by the GCN student in order to better understand the properties of our \med. 
We follow \cite{dgi} and focus our analysis exclusively on Cora \cite{DBLP:journals/ir/McCallumNRS00,DBLP:conf/iclr/BojchevskiG18} because it has the smallest number of nodes among the node-level benchmarks, which notably aids clarity.

Figure \ref{fig:tsne} gives a standard set of ``evolving'' t-SNE plots of the embeddings learnt by three models, namely the vanilla GCN, the student GCN trained with our \med, and the teacher GCNII, respectively. 
For all the three subfigures, we can observe that the learnt embeddings’ 2D projections exhibit discernible clustering in the projected space, which corresponds to the seven topic classes of Cora. 
We further notice that the 7-category scientific papers can be differentiated more effectively by student GCN equipped with \med~than by a vanilla GCN. 

Figure \ref{fig:tsne} qualitatively shows that \med~not only increases the accuracy of classification, but also enables the student GCN to learn high-quality node representations.
To obtain a more accurate and convincing conclusion, we calculate Silhouette scores \cite{rousseeuw1987silhouettes} for the three projections.
Specifically, the Silhouette score \cite{rousseeuw1987silhouettes} of the  embeddings learned by student GCN is 0.2638, which compares favorably with the score of 0.2196 for the vanilla GCN.


%% file: section/related.tex
\section{Related Work} 
In this part, we first introduce existing work that adapts knowledge distillation to graph domains in Section \ref{sec:kd4gnn}. Next, we review applications of adversarial training in graph domains in Section \ref{sec:adv}.

\subsection{Knowledge Distillation for Graph Models} \label{sec:kd4gnn}
Knowledge distillation has achieved great success for network compression in visual learning and language modeling tasks \cite{DBLP:conf/cvpr/BergmannFSS20,fitnets, tinybert,distilbert}.
However, directly applying the established approaches in visual learning and language modeling to graph domains is not applicable as graphs contain both features and topological structures.

Recent success in GNNs impels the advent of knowledge distillation for GNNs.
Among existing work, \cite{DBLP:conf/cvpr/YangQSTW20,DBLP:conf/sigmod/ZhangMSJCR020,DBLP:conf/www/0002LS21} are the most relevant to this paper as they all use the teacher-student architecture and focus on node classification.
However, the adaptive distillation strategy makes our work distinct from existing research \cite{DBLP:conf/cvpr/YangQSTW20,DBLP:conf/sigmod/ZhangMSJCR020,DBLP:conf/www/0002LS21}.
Specifically, LSP \cite{DBLP:conf/cvpr/YangQSTW20} aligns node representations of teacher and student with kernel function based distance;
\citet{DBLP:conf/sigmod/ZhangMSJCR020} and \citet{DBLP:conf/www/0002LS21} use Euclidean distance to match the probability distributions of teacher and student.
Contrary to them, our proposed \med~adversarially trains a discriminator and a generator to adaptively detect and decrease the discrepancy between teacher and student.
Moreover, \cite{DBLP:conf/cvpr/YangQSTW20,DBLP:conf/sigmod/ZhangMSJCR020,DBLP:conf/www/0002LS21} merely conduct node-level classification and focus on graphs with less than 100K nodes, while the proposed \med~is widely applicable to both node-level and graph-level classification tasks and performs well on graphs with number of nodes varying from 2K to 2M.

\subsection{Adversarial Training for Graph Models} \label{sec:adv}
The idea of adversarial training originates from generative adversarial networks (GANs) \cite{gan}, where the generator and discriminator compete with each other to improve their performance.

In recent years, adversarial training has demonstrated superior performance in graph domains for different aims.
Specifically, \citet{wang2018graphgan} and \citet{feng2019graph} leverage the adversarial architecture to learn universal and robust graph representations, while \citet{dai2018adversarial} explore adversarial attack on graph structured data.
\citet{alam2018domain} perform adversarial domain adaptation with graph models, while \citet{suresh2021adversarial} develop adversarial graph augmentation to improve the performance of self-supervised learning.
Different from them, this work aims to conduct adversarial knowledge distillation for graph models, which leads to the contribution.

%% file: section/conclusion.tex
\section{Conclusion}
Over-stacked GNNs are usually expressive and powerful on large-scale graph data.
To compress deep GNNs, we present a novel adversarial knowledge distillation framework in graph domains, namely \med, which introduces adversarial training to topology-aware knowledge transfer for the first time.
By adversarially training a discriminator and a generator, \med~is able to transfer both inter-node and inter-class correlations from a complicated teacher GNN to a compact student GNN (i.e., the generator).
Experiments demonstrate that \med~yields consistent and significant improvements across node-level and graph-level tasks on ten benchmark datasets.
The student GNNs trained this way achieve competitive or even superior results to their teacher graph models, 
while requiring only a small proportion of parameters.
In the future work, we plan to explore the potential application of \med~on graph tasks beyond classification.


%% file: section/appendix.tex
\section{Code for Graph Classification}\label{sec:alg2}
\begin{algorithm}[!ht]
    \caption{\med~for graph-level classification.} \label{alg:graph-level}
    \begin{algorithmic}[1]
        \Require Graphs $\{\mathcal{G}_1, \cdots, \mathcal{G}_N\}$, and the pretrained teacher $G^T$.
        \Ensure The learnt student model $G^S$.
        \While{not converge} 
            \ForAll{graph $\mathcal{G}\in \{\mathcal{G}_1, \cdots, \mathcal{G}_N\}$}
                \State $\mathbf{H}^T, \mathbf{z}_\mathcal{G}^T=G^T(\mathbf{X}_\mathcal{G}, \mathbf{A}_\mathcal{G})$; $\mathbf{s}_\mathcal{G}^T=\frac{1}{|\mathcal{V}|}\sum_{v\in \mathcal{V}}\mathbf{h}_v^T$
                \State $\mathbf{H}^S, \mathbf{z}_\mathcal{G}^S=G^S(\mathbf{X}_\mathcal{G}, \mathbf{A}_\mathcal{G})$; $\mathbf{s}_\mathcal{G}^S=\frac{1}{|\mathcal{V}|}\sum_{v\in \mathcal{V}}\mathbf{h}_v^S$
                \ForAll{node $v\in \mathcal{V}\subset \mathcal{G}$}
                    \State Update $D_e$ to distinguish $(\mathbf{h}_v^T, \mathbf{s}_\mathcal{G}^T)$ and $(\mathbf{h}_v^S, \mathbf{s}_\mathcal{G}^T)$
                    \State Update $D_e$ to distinguish $(\mathbf{h}_v^T, \mathbf{s}_\mathcal{G}^S)$ and $(\mathbf{h}_v^S, \mathbf{s}_\mathcal{G}^S)$
                    \For{node $u\in \mathcal{N}(v)$}
                        \State Update $D_e$ to distinguish $(\mathbf{h}_v^T, \mathbf{h}_u^T)$ and $(\mathbf{h}_v^S, \mathbf{h}_u^S)$
                        \State Update the parameters of $G^S$ to fool $D_e$ via Eq.~\ref{eq:emb_D}
                    \EndFor
                \EndFor
                \State Update $D_\ell$ to distinguish $\mathbf{z}_\mathcal{G}^T$ and $\mathbf{z}_\mathcal{G}^S$
                \State Update the parameters of $G^S$ to fool $D_\ell$ via Eq.~\ref{eq:logits_D}
            \EndFor
        \EndWhile 
        \State \Return $G^S$
    \end{algorithmic}
\end{algorithm}

\section{Datasets} \label{app:dataset}
We detail all datasets as follows.
\begin{itemize}[leftmargin=*]
    \item \textbf{Cora} \cite{DBLP:journals/ir/McCallumNRS00,DBLP:conf/iclr/BojchevskiG18} and \textbf{CiteSeer} \cite{DBLP:journals/aim/SenNBGGE08} are networks of computer science publications. Each node in the two networks represents a publication and each directed edge means a citation. Each node is annotated with a vector of binary word indicators and a label indicating the paper topic.
    \item \textbf{PubMed} \cite{namata2012query} is a set of articles related to diabetes from the PubMed database. Node features are TF/IDF-weighted word frequencies and the labels specify the type of diabetes.
    \item \textbf{Flickr} \cite{DBLP:conf/eccv/McAuleyL12,graphsaint} is an undirected graph of images. Nodes are \emph{images}, and edges indicate the connected two images share some common properties (e.g., geographic location, gallery, and users commented, etc.).
    Node features are the 500-dimensional bag-of-words representation of the images.
    We adopt the labels and dataset split in \cite{graphsaint}.
    \item \textbf{Arxiv} \cite{ogb} is a directed graph that represents the citation network between all computer science arXiv papers indexed by MAG \cite{mag}. 
    Node features are the averaged skip-gram word embeddings of the paper title and abstract.
    \item \textbf{Reddit} \cite{reddit,graphsaint} is an undirected graph constructed from an online discussion forum. Nodes are posts belonging to different communities and each edge indicates the connected posts are commented by the same user.
    We use the sparse version of Reddit dataset, which contains about 23M edges instead of more than 114M edges \cite{graphsaint, gas}. Besides, we follow the same inductive setting in \cite{reddit, graphsaint, gas}, i.e., we do not require all nodes in the graph are present during training.
    \item \textbf{Yelp} \cite{graphsaint} is constructed with the data of business, users and reviews provided in the open challenge website. Each node is a customer and each edge implies the connected users are friends. Node features are Word2Vec embeddings of the user's reviews. We follow \cite{graphsaint, gas} to use the categories of the businesses reviewed by a user as the multi-class label of the corresponding node.
    \item \textbf{Products} \cite{ogb} is an undirected graph that represents an Amazon product co-purchasing network. Nodes are products sold on Amazon, and edges indicate that the connected products are purchased together. Node features are bag-of-words vectors of the product descriptions, and node labels are categories of the products.
    \item \textbf{Molhiv} \cite{ogb} and \textbf{Molpcba} \cite{ogb} are two molecular property prediction datasets of different sizes. Each graph represents a molecule, where nodes are atoms, and edge are chemical bonds. Edge features indicate bond type, bond stereochemistry, and whether the bond is conjugated. Both node features and edge features are considered to predict the target molecular properties as accurately as possible.
\end{itemize}

\section{Experimental Details} \label{app:detail}

\subsection{Implementation Details.}
We implement \med~in PyTorch and run it on a single NVIDIA GeForce RTX 2080Ti graphics card.
Our implementation generally follows the open source codebases of deep graph library. 
We instantiate \med~with a generator-discriminator architecture.
For the generator part, we employ almost the same experimental settings including initialization, optimization and hyper-parameters as the corresponding vanilla student GNN.
For convenience, we set the generator's node embedding dimension to be the same as the teacher model's embedding dimension.
For the discriminator part, $D_\ell$ and $D_e$ are uniformly initialized and all-one initialized, respectively. 
We sum the adversarial losses produced by $D_\ell$ and $D_e$ without any tuned weight.
We train the discriminator using Adam optimizer \cite{adam} with a learning rate varying from 0.05 to 0.001.
\med~updates the parameters of generator and discriminator with a ratio of $k:1$, which implies that the discriminator is update once after the generator is updated k times. 
We perform grid search to find a suitable $k$ among $\{1, 5, 10, 20, 30\}$ for each dataset.

\subsection{Model selection}

We detail the teacher graph models as follows.
\begin{itemize}[leftmargin=*]
    \item \textbf{GCNII} \cite{gcn2} is an extension of the vanilla GCN \cite{gcn}. \citet{gcn2} propose two effective techniques---namely initial residual and identity mapping---to deepen the graph convolution layers.
    GCNII increases the number of graph convolution layers from 2 to 64, while the performance on node classification is not affected by the over-smoothing issue \cite{DBLP:conf/aaai/LiHW18,jknet}.
    \item \textbf{GAMLP} \cite{gamlp} is a powerful and over-parameterized graph learning model based on the reception field attention. Specifically, GAMLP \cite{gamlp} incorporates three principled attention mechanisms---namely smoothing attention, recursive attention, and jumping knowledge (JK) attention---into the representation learning process.
    \item \textbf{HIG} \footnote{\url{https://github.com/TencentYoutuResearch/HIG-GraphClassification}.} is proposed as a node augmentation method to solve the graph classification task. HIG randomly selects nodes and applies heterogeneous interpolation, then adds KL divergence constraint loss to make the distributions of augmented features be similar. For Molhiv \cite{ogb}, HIG built upon DeeperGCN \cite{li2020deepergcn} achieves state-of-the-art performance. For Molpcba \cite{ogb}, HIG that selects Graphormer \cite{ying2021transformers} as the backbone achieves state-of-the-art performance.
\end{itemize}

We detail the student graph models as follows.
\begin{itemize}[leftmargin=*]
    \item \textbf{GCN} \cite{gcn} simplifies graph convolutions by stacking layers of first-order Chebyshev polynomial filters. It has been proved to be one of the most popular baseline GNN architectures.
    \item \textbf{Cluster-GCN} \cite{cluster-gcn} relieves the out-of-memory issue for GCN when scaling to large-scale graphs. Specifically, Cluster-GCN designs node batches based on efficient graph clustering algorithms, which leads to great computational benefits.
    \item \textbf{GIN} \cite{gin} generalizes the WL test and provably achieves great discriminative power among GNNs. Based on the theory of ``deep multisets'', GIN learns to embed the subtrees in WL test to the low-dimensional space. By this means, GIN is able to discriminate different structures, and capture dependencies between graph structures as well.
\end{itemize}